# Orthogonal Stochastic Configuration Networks with Adaptive Construction Parameter for Data Analytics

Wei Dai, *Senior Member, IEEE*, Chuanfeng Ning, Shiyu Pei, Song Zhu, and Xuesong Wang, *Member, IEEE*

**Abstract**—As a randomized learner model, SCNs are remarkable that the random weights and biases are assigned employing a supervisory mechanism to ensure universal approximation and fast learning. However, the randomness makes SCNs more likely to generate approximate linear correlative nodes that are redundant and low quality, thereby resulting in non-compact network structure. In the light of a fundamental principle in machine learning, that is, a model with fewer parameters holds improved generalization. This paper proposes orthogonal SCN, termed OSCN, to filtrate out the low-quality hidden nodes for network structure reduction by incorporating Gram-Schmidt orthogonalization technology. The universal approximation property of OSCN and an adaptive setting for the key construction parameters have been presented in details. In addition, an incremental updating scheme is developed to dynamically determine the output weights, contributing to improved computational efficiency. Finally, experimental results on two numerical examples and several real-world regression and classification datasets substantiate the effectiveness and feasibility of the proposed approach.

## 1 Introduction

NEURAL networks (NNs) are an essential part of machine learning, because they can propose efficient solutions to complex data analysis tasks that are difficult to be solved by conventional methods [1]-[4]. Feedforward neural networks (FNNs), the kind of classical NNs, have been widespread used and studied due to its simple construction and strong nonlinear mapping ability [5], [6]. However, the generalization of FNNs is very sensitive to the network parameter settings, such as learning rate, owing to the use of gradient descent algorithms for training [7], [8]. Similarly, this training approach can be subjected to local minima, time-consuming problems, and some other limitations [9], [10].

Randomized algorithms have shown great potential in exploring fast learning and low computational cost [11]-[14]. Therefore, random vector functional link networks (RVFLNs), a kind of single hidden layer FNNs with randomized algorithms, are presented [15]-[21]. In RVFLNs, the input weights and biases are randomly assign from certain fixed intervals range and remain constant. And, the output weights are obtained by solving a linear equation [22]. Although RVFLNs have demonstrated significant potential, it is difficult to construct an appropriate network structure to accomplish modeling tasks. In general, it is challenging, if not impossible, to obtain a proper network topology via human experience. The network with too large or too small size will suffer from performance degrading.

Constructive algorithm starts with a simple network and gradually adds hidden nodes (hidden nodes and weights) until a predefined condition could be satisfied [23], [24]. This construction feature makes it possible for the constructive algorithm to find the most suitable network structure. Further, RVFLNs with the constructive algorithms is proposed, called (IRVFLNs). However, recent work in [25] indicates that these IRVFLN-based models have difficulties in guaranteeing the universal approximation property, as a consequence of extensive scope setting lacking scientific justification. In [26], the poor approximation performance of common RVFLNs with the fixed parameter scope is explained in more detail.

According to the previous work, an advanced randomized learner model, known as stochastic configuration networks

This work was supported in part by the National Natural Science Foundation of China under Grant 61973306, in part by the Natural Science Foundation of Jiangsu Province under Grant BK20200086, in part by the Open Project Foundation of State Key Laboratory of Synthetical Automation for Process Industries under Grant 2020-KF-21-10, and in part by the Postgraduate Research & Practice Innovation Program of Jiangsu Province under Grant KYCX22_2560. *(Corresponding author: Wei Dai)*

Wei Dai is with the School of Information and Control Engineering and Engineering Research Center of Intelligent Control for Underground Space, Ministry of Education, China University of Mining and Technology, Xuzhou 221116, China, and also with State Key Laboratory of Synthetical Automation for Process Industries, Northeastern University, Shenyang 110819, Liaoning, China (e-mail: daiwei_neu@126.com; weidai@cumt.edu.cn).

Chuanfeng Ning, Shiyu Pei and Xuesong Wang are with the School of Information and Control Engineering, China University of Mining and Technology, Xuzhou 221116, China and also with Engineering Research Center of Intelligent Control for Underground Space, Ministry of Education, China University of Mining and Technology, Xuzhou, 221116, China (e-mail: TS20060090A31@cumt.edu.cn;psy_cumt@126.com;wangxuesongcumt@163.com).

Song Zhu is with the School of Mathematics, China University of Mining and Technology, Xuzhou 221116, China (e-mail: songzhu@cumt.edu.cn).

(SCNs) was reported in [27]. Specifically, SCNs employ an incremental construction approach where a hidden node and all its connected weights and biases are added in each iteration. Also, SCN takes advantage of a scope setting vector to select a set of candidate weights and biases randomly, under the constraints of a supervisory mechanism. It is this step that make SCNs and its variants, including deep version [28], [29], robust version [30], ensemble version [31] and 2D version [32] exhibit the satisfactory performance in big data, uncertain data problems and image data modelling tasks. However, SCNs are more likely to generate approximate linear correlative nodes resulting from the randomness, even if the supervisory mechanism is employed. These nodes with small outputs are redundant and low quality, which easily give rise to ill-conditioned hidden output matrix, depreciating generalization performance. Simultaneously, the redundancy among a myriad of candidate nodes induces large model size, thus contributing little to more compact network structure.

Focusing on the abovementioned problems, the improved SCNs, termed as orthogonal SCN (OSCN) is proposed. The Gram-Schmidt orthogonalization technology is integrated into SCNs to evaluate the level of correlation among random generated nodes, which filters out redundant nodes and achieves better performance. This paper proposes OSCN under the following contributions and novelties.

1) The Gram-Schmidt orthogonalization technology is adopted to evaluate and filter out low-quality candidate nodes in the stochastic configuration process, thereby simplifying the structure network and enhancing generalization performance.
2) In the orthogonal framework, the optimal output weight can be determined by taking advantage of a constructive scheme, which avoids complicated and time-consuming retraining procedure and results in high computational efficiency to a certain extent.
3) The universal approximation property of OSCN is established in the form of orthogonal supervisory mechanism. Additionally, an adaptive setting for construction parameters, which can be adaptively generated in the supervisory mechanism, is given.

The rest of this paper is constructed as follows. We revisit SCNs in Section 2. Section 3 introduces OSCN in terms of theoretical analysis and algorithm implementation. In Section 4, comparative experimental results and analysis are shown. Finally, section 5 concludes and indicates the future work.

## 2 Brief Reviews of SCNs

SCNs, as a class of advanced universal approximators, have demonstrated its superiority in a wide range of applications, attributing to fast learning speed and sound generalization [27].

Assuming that we have built a network model with $L$-1 hidden nodes: $f_{L-1}(x) = \sum_{j=1}^{L-1} \beta_j g_j(w_j^T x + b)$. $e_{L-1} = f - f_{L-1}$ represents the residual error. If the current residual error fails to achieve the predefined condition. SCNs will incrementally produce $L$-th hidden nodes associated with a set of candidate parameters giving rise to the current error approach to the predefined condition.

The algorithm implementation of SCN can be expressed as follows:

● Initialization

Give $X=\{x_1, x_2,…, x_N\}$ to be the $N$ inputs to a training dataset, where $x_i \in \mathbb{R}^d$. And accordingly, give $T=\{t_1, t_2,…,t_N\}$ to be $N$ outputs, where $t_i \in \mathbb{R}^m$. Relative parameters in the incremental constructive process, which can refer to **OSCN Algorithm** as described using pseudo codes, are illustrated in details.

● Hidden Parameter Configuration

Assigning $w_L$ and $b_L$ stochastically from the support scope to in acquisition of a set of candidates random basis function $h_L = g_L(w_L^T x + b_L)(0 < \|h\| \le b_g)$, which satisfies the following inequality:

$$\langle e_{L-1,q}, h_L \rangle^2 \ge b_g^2(1-r-\mu_L)\|e_{L-1,q}\|, q = 1, 2, \cdots, m \quad (1)$$

where $r$ and $\mu_L$ are contractive parameters.

● Output Weight Determination

There are three original schemes, including SC-I, SC-II, and SC-III, for evaluating the output weights. Three algorithmic implementations are illustrated in Fig. 1. Concretely, SC-I updates the output weights of the newly added hidden node, and removes the necessity for recalculating the former. SC-II recalibrates a portion of the existing output weights according to predefined sliding window size and achieves the suboptimal solution to the output weights. The output weights of all existing hidden nodes in SC-III are assigned by means of solving a global optimal problem, which can be more likely to achieve effectively in targeting on a universal approximator during the incremental learning process.
- Calculate the current residual error $e_L$ and update $e_0:=e_L$, update $L:=L+1$ until the network meets the predefined conditions: $L \leq L_{max}$ or $\|e_0\| > \varepsilon$.

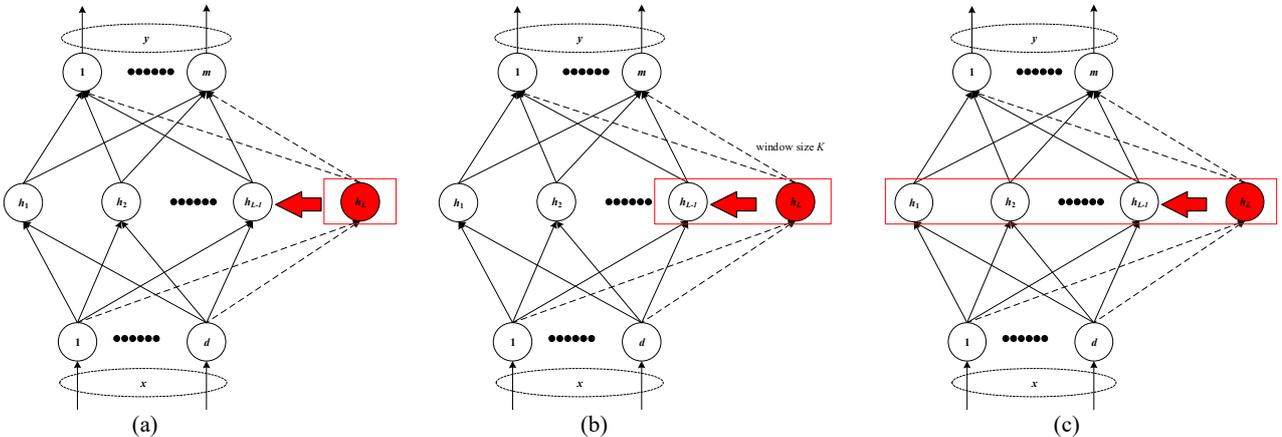

Fig. 1. Algorithmic implementations of SCNs. (a) SC-I. (b) SC-II. (c) SC-III.

**Remark 1:** SC-III outperforms the others (SC-I, SC-II) in terms of generalization and convergence, but suffers form largest computation load due to Moore-Penrose generalized inverse [33] implemented. The newly added output weights in SC-I are determined employing a constructive scheme, which contributes to minimal computation load, but involves worst convergence. From the perspective of both computation load and convergence, SC-II compromises in comparison of SC-I, SC-III.

**Remark 2:** Note that the fastest decrease in residual error depends on construction parameters $(r, \mu_L)$, as shown in (1). Hence, how to select appropriate construction parameters is directly making a difference to model learning. Besides, as the construction process proceeds, the newly added hidden nodes with smaller outputs due to the randomness are less conducive to the reduction of residual error. Even though these nodes may be selected to add the existing network, they are less likely to maintain the network compactness and achieve faster convergence speed. Therefore, the quality of nodes requires to be ameliorated.

To mitigate the weakness mentioned above, an orthogonal version of SCN, termed OSCN, which can be built efficiently with high-quality nodes, and achieve the global optimal parameters, is proposed.

## 3 ORTHOGONAL STOCHASTIC CONFIGURATION NETWORKS

In this section, the proposed OSCN is detailed. Firstly, the description of our proposed model is presented, followed by theoretical analysis. Afterward, we give the overall procedure for OSCN in **OSCN Algorithm**.

### A. Description of OSCN model

The OSCN framework can be summarized in Fig. 2. This process can be generalized as configuring random parameters associated with the first hidden nodes first, then the subsequent hidden nodes are made orthogonal to each other to guarantee that the networks will converge more efficiently without redundant nodes. Details of constructing OSCN are outlined below.

Give $X = \{x_1, x_2, \ldots, x_N\}$ to be the $N$ inputs to a training dataset, $x_i = [x_{i,1}, x_{i,2}, \ldots, x_{i,d}] \in \mathbb{R}^d$. And accordingly, give $T = \{t_1, t_2, \ldots, t_N\}$ to be $N$ outputs, $t_i = [t_{i,1}, t_{i,2}, \ldots, t_{i,m}] \in \mathbb{R}^m$. Suppose that OSCN has already constructed $L$-1 hidden nodes, let the candidate nodes in stochastic configuration of $L$th hidden node can be written as follows:

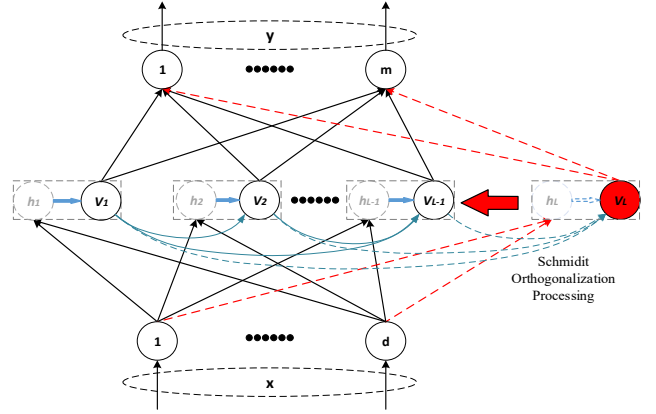

Fig. 2. Schematic diagram of orthogonal method.

$$h_L = [g_L(w_L^T x_1 + b_L), g_L(w_L^T x_2 + b_L), \cdots, g_L(w_L^T x_N + b_L)]^T \quad (2)$$

Considering the randomness, we introduce the Gram-Schmidt into stochastic configuration process to guarantee the quality of candidate nodes from the perspective of collinearity [34]. Then the orthogonal vector of the candidate node can be calculated by

$$v_L = \begin{cases} h_1, L = 1 \\ h_L - \frac{\langle v_1, h_L \rangle}{\langle v_1, v_1 \rangle} v_1 - \frac{\langle v_2, h_L \rangle}{\langle v_2, v_2 \rangle} v_2 - \cdots - \frac{\langle v_{L-1}, h_L \rangle}{\langle v_{L-1}, v_{L-1} \rangle} v_{L-1}, L \neq 1 \end{cases} \quad (3)$$

To avoid generating approximate linear correlative hidden nodes with small output weights that are inefficient on decrease in residual error, a small positive number $\sigma$ is given to estimate whether the candidate node is redundant for residual error reduction, if $\|v_L\| \geq \sigma$, that means this one can be considered as a good one. The best-hidden node added to the network can be configured by maximizing the supervisory mechanism among a multitude of candidate nodes. After orthogonalization $\text{span}\{v_1, v_2, \ldots, v_L\} = \text{span}\{h_1, h_2, \ldots, h_L\}$ which means $v_1, v_2, \ldots, v_L$ are equivalent to $h_1, h_2, \ldots, h_L$. Therefore, the OSCN model can be formulated as $f_L = f_{L-1} + v_L \beta_L$, the current residual error $e_{L-1}$ is denoted by $e_{L-1} = f - f_{L-1} = [e_{L-1,1}, e_{L-1,2}, \ldots, e_{L-1,m}] \in \mathbb{R}^{N \times m}$ and the output weights can be expressed as $\beta = [\beta_1, \beta_2, \ldots, \beta_L]^T$, where $\beta_L = [\beta_{L,1}, \beta_{L,2}, \ldots, \beta_{L,m}] \in \mathbb{R}^{1 \times m}$.

### B. Output weight evaluation

Although OSCN can improve the quality of candidate nodes to help in building compact network for better performance through filtering out redundant nodes, it may take a little bit more training time to each orthogonalization during $T_{max}$ stochastic configuration. Fortunately, benefitting from the orthogonalization construction, the proposed OSCN model can update the output weights similar to SC-I, and its convergence performance is similar to SC-III. In the next part, we will prove this nature.

In the orthogonal framework, the output weights are

analytically determined by

$$\beta_{L,q} = \frac{\langle e_{L-1,q}, v_L \rangle}{\langle v_L, v_L \rangle}, q = 1, 2, \ldots, m \qquad (4)$$

Notice that, for an OSCN with $L$ hidden nodes, we have $f_L = f_{L-1} + v_L \beta_L$ and $<v_i, v_j> = 0, i \neq j$ so that

$$\begin{aligned} e_L &= f - f_L \\ &= f - (f_{L-1} + v_L \beta_L) \\ &= e_{L-1} - v_L \beta_L \end{aligned} \qquad (5)$$

Thus, for $e_i = [e_{i,1}, e_{i,2}, \ldots, e_{i,m}], i = 1, 2, \cdots, L$, according to Eq. (5), we have

$$e_{i,q} = e_{i-1,q} - v_i \beta_{i,q}, q = 1, 2, \cdots, m. \qquad (6)$$

Substituting Eq. (4) into Eq. (6), it can be known that,

$$\begin{aligned} \langle e_{i,q}, v_i \rangle &= \langle e_{i-1,q} - v_i \beta_{i,q}, v_i \rangle \\ &= \langle e_{i-1,q}, v_i \rangle - \langle v_i \beta_{i,q}, v_i \rangle \\ &= \langle e_{i-1,q}, v_i \rangle - \beta_{i,q} \langle v_i, v_i \rangle \\ &= \langle e_{i-1,q}, v_i \rangle - \frac{\langle e_{i-1,q}, v_i \rangle}{\langle v_i, v_i \rangle} \langle v_i, v_i \rangle \\ &= 0 \end{aligned} \qquad (7)$$

Then,

$$\langle e_i, v_i \rangle = e_i^T v_i = \begin{bmatrix} e_{i,1}^T \\ e_{i,2}^T \\ \vdots \\ e_{i,m}^T \end{bmatrix} v_i = \begin{bmatrix} e_{i,1}^T v_i \\ e_{i,2}^T v_i \\ \vdots \\ e_{i,m}^T v_i \end{bmatrix} = \begin{bmatrix} \langle e_{i,1}, v_i \rangle \\ \langle e_{i,2}, v_i \rangle \\ \vdots \\ \langle e_{i,m}, v_i \rangle \end{bmatrix} = 0 \qquad (8)$$

So we can get

$$\langle e_1, v_1 \rangle = 0 \qquad (9)$$

$$\langle e_2, v_2 \rangle = 0 \qquad (10)$$

$$\begin{aligned} \langle e_2, v_1 \rangle &= \langle e_1 - v_2 \beta_2, v_1 \rangle \\ &= \langle e_1, v_1 \rangle - \beta_2^T v_2^T v_1 \\ &= \langle e_1, v_1 \rangle - \beta_2^T \langle v_2, v_1 \rangle \\ &= 0 \end{aligned} \qquad (11)$$

The above equations can be summarized as $e_1 \perp \text{span}\{v_1\}$; $e_2 \perp \text{span}\{v_1, v_2\}$; Suppose for all $2 \leq k \leq L$, $e_{k-1} \perp \text{span}\{v_1, v_2, \cdots, v_{k-1}\}$; According to $<v_k, v_j> = 0, k \neq j$ and Eq. (8), $<e_k, v_k> = 0$. For all $1 \leq j \leq k-1$,

$$\begin{aligned} \langle e_k, v_j \rangle &= \langle e_{k-1} - v_k \beta_k, v_j \rangle \\ &= \langle e_{k-1}, v_j \rangle - \beta_k^T \langle v_k, v_j \rangle \\ &= 0 \end{aligned} \qquad (12)$$

So $e_k \perp \text{span}\{v_1, v_2, \cdots, v_k\}$, that is, $e_L \perp \text{span}\{v_1, v_2, \cdots, v_L\}$.

For the least squares solution $\beta^* = \underset{\beta}{\arg\min} \|T - V_L \beta\|$, $\beta^* \in R^{L \times m}$, given the deduction above, we have

$$\langle e_L, V_L \rangle = e_L^T V_L (\beta - \beta^*) = 0 \qquad (13)$$

where $V_L = [v_1, v_2, \cdots, v_L]$. Thus,

$$\begin{aligned} \|T - V_L \beta^*\|^2 &= \|T - V_L \beta + V_L \beta - V_L \beta^*\|^2 \\ &= \|T - V_L \beta + V_L (\beta - \beta^*)\|^2 \\ &= \|T - V_L \beta\|^2 + \|V_L (\beta - \beta^*)\|^2 \\ &\quad + 2 \langle e_L, V_L (\beta - \beta^*) \rangle \\ &= \|T - V_L \beta\|^2 + \|V_L (\beta - \beta^*)\|^2 \\ &\geq \|T - V_L \beta\|^2 \end{aligned} \qquad (14)$$

$\|T - V_L \beta^*\| = \|T - V_L \beta\|$ holds if and only if $\beta = \beta^*$. Therefore, $\beta$ based on Eq. (4) is also the LS solution of $\|T - V_L \beta\| = 0$.

As a consequence, OSCN can train the newly added output weight while maintaining the same effect as global method that requires calculating the output weights all together after node added. In this way, OSCN can be more likely to avoid the complicated and time-consuming retraining procedure and make up for some of the orthogonal computation time to some extent.

### C. Universal approximation property

The theoretical analysis is investigated on the universal approximation property, which can serve as an extension of that given in [27].

OSCNs with $L$-1 hidden nodes have been constructed: $f_{L-1}(x) = \sum_{j=1}^{L-1} v_j \beta_j$, $e_{L-1} = f - f_{L-1} = [e_{L-1,1}, e_{L-1,2}, \cdots, e_{L-1,m}]$ where $v_i$ represents the $i$th hidden output after orthogonalization. Represent the current residual error as $e_L = e_{L-1} - v_L \beta_L$, $\beta_L = [\beta_{L,1}, \beta_{L,2}, \cdots, \beta_{L,q}, \cdots, \beta_{L,m}]$.

**Theorem 1.** Suppose that span ($\Gamma$) is dense in $L_2$ space. Given $0 < r < 1$ and a non-negative real number sequence $\{\mu_L\}$ with $\lim_{L \to +\infty} \mu_L = 0$ and $\mu_L \leq 1 - r$. For $L = 1, 2, \cdots$ denoted by

$$\delta_L = \sum_{q=1}^{m} \delta_{L,q}, \delta_{L,q} = (1 - r - \mu_L) \|e_{L-1,q}\|^2 \qquad (15)$$

There exists $V_L = [v_1, v_2, \cdots, v_L]$ such that span$\{h_1, h_2, \cdots, h_L\}$ = span$\{v_1, v_2, \cdots, v_L\}$ concentrating on satisfying the following orthogonal form of inequality constraints (supervisory

mechanism):

$$\langle e_{L-1,q}, v_L \rangle^2 \geq v_L^2 \delta_{L,q}, q=1,2,\cdots,m \quad (16)$$

Then, the output weights can be obtained by

$$\beta_{L,q} = \frac{\langle e_{L-1,q}, v_L \rangle}{\langle v_L, v_L \rangle}, q=1,2,\cdots,m \quad (17)$$

Then, we have $\lim_{L \to +\infty} \|f - f_L\| = 0$.

For the purpose of simplicity, a set of instrumental variables $\xi_L = \sum_{q=1}^{m} \xi_{L,q}$ are introduced as follows:

$$\xi_{L,q} = \frac{\left(e_{L-1,q}^{\mathrm{T}} \cdot v_L\right)^2}{v_L^{\mathrm{T}} \cdot v_L} - (1 - r - \mu_L) e_{L-1,q}^{\mathrm{T}} \cdot e_{L-1,q} \quad (18)$$

**Proof.** According to Eq. (17), the verification of OSCN is similar to SCNs, it is easy to validate that the sequence $\|e_L\|^2$ is monotonically decreasing and converges.

From Eq. (15)–(17), we can further obtain

$$\begin{aligned}
\|e_L\|^2 - (r + \mu_L)\|e_{L-1}\|^2 &= \sum_{q=1}^{m} \langle e_{L-1,q} - v_L \beta_{L,q}, e_{L-1,q} - v_L \beta_{L,q} \rangle \\
&\quad - \sum_{q=1}^{m} (r + \mu_L) \langle e_{L-1,q}, e_{L-1,q} \rangle \\
&= (1 - r - \mu_L)\|e_{L-1}\|^2 - \frac{\sum_{q=1}^{m} \langle e_{L-1,q}, v_L \rangle^2}{\|v_L\|^2} \\
&= \delta_L - \frac{\sum_{q=1}^{m} \langle e_{L-1,q}, v_L \rangle^2}{\|v_L\|^2} \\
&\leq 0
\end{aligned} \quad (19)$$

Therefore, $\|e_L\|^2 - (r + \mu_L)\|e_{L-1}\|^2 \leq 0$. It is worthy of mentioning that $\lim_{L \to \infty} \mu_L \|e_{L-1}\|^2 = 0$ while $\lim_{L \to \infty} \mu_L = 0$. According to the abovementioned equations, we can easily get that $\lim_{L \to +\infty} \|e_{L-1}\|^2 = 0$, that is $\lim_{L \to +\infty} \|e_L\| = 0$. Above completes the whole process of proof. This completes the proof.

*D. Adaptive Construction Parameter*

**Theorem 1** provides inequality constraints (supervisory mechanism) to guarantee the universal approximation property. It can be easily observed that the final hidden output $v_L$ added to the network can be configured through selecting the one that maximizes $\xi_L$ among a collection of candidate nodes. From Eq. (18), it can be found that the construction parameters $r$ and $\mu_L$ are also the key factors making a difference to the candidate set. In SCNs, $r$ is an incremental sequence within an adjustable interval (0.9-1), and $\mu_L = (1-r)/(L+1)$. Although $r$ is kept unchanged in the incremental constructive process, this artificial setting for $r$ may abandon a plethora of weights and biases selected randomly over some intervals imposed restrictions on a range. Consequently, the assignment of candidate random parameters is more likely to being confronted with time-consuming problems or even more unnecessary fails.

In **Theorem 2**, an adaptive setting for construction parameters is provided.

**Theorem 2.** Given a non-negative sequence $\tau_L = r + \mu_L = (L/(L+1) + 1/(L+1)^2)$, we have $\|e_L\|^2 \leq \tau_L \|e_{L-1}\|^2$ and $\lim_{L \to \infty} \|e_L\| = 0$.

**Proof.** It is easy to obtain $\lim_{L \to +\infty} \tau_L = 1$ The theoretical result stated in **Theorem 1**, we can get

$$\begin{aligned}
\|e_L\|^2 &\leq \tau_L \|e_{L-1}\|^2 \\
&\leq \prod_{j=1}^{L} \tau_j \|e_0\|^2 \\
&\leq \prod_{j=1}^{L} \left(\frac{j}{j+1} + \frac{1}{(j+1)^2}\right) \|e_0\|^2 \\
&= \prod_{j=1}^{L} \left(1 - \frac{1}{j+1}\left(1 - \frac{1}{j+1}\right)\right) \|e_0\|^2
\end{aligned} \quad (20)$$

Similar to $1 - x < e^{-x}, x > 0$, we have:

$$\begin{aligned}
&\prod_{j=1}^{L} \left(1 - \frac{1}{j+1}\left(1 - \frac{1}{j+1}\right)\right) \|e_0\|^2 \\
&< \exp\left(-\sum_{j=1}^{L} \frac{1}{j+1}\left(1 - \frac{1}{j+1}\right)\right) \|e_0\|^2 \\
&= \exp\left(-\sum_{j=1}^{L} \frac{1}{j+1} + \sum_{j=1}^{L} \frac{1}{(j+1)^2}\right) \|e_0\|^2
\end{aligned} \quad (21)$$

Then, considering

$$\begin{aligned}
\sum_{j=1}^{L} \frac{1}{j+1} &> \ln\left(1 + \frac{1}{2}\right) + \ln\left(1 + \frac{1}{3}\right) + \ldots + \ln\left(1 + \frac{1}{L+1}\right) \\
&= \ln\left(\left(\frac{3}{2}\right) \times \left(\frac{4}{3}\right) \times \ldots \times \left(\frac{L+2}{L+1}\right)\right) \\
&= \ln\left(1 + \frac{L}{2}\right)
\end{aligned} \quad (22)$$

and

$$\begin{aligned}
\sum_{j=1}^{L} \frac{1}{(j+1)^2} &= \frac{1}{2 \times 2} + \frac{1}{3 \times 3} + \ldots + \frac{1}{(L+1) \times (L+1)} \\
&< \frac{1}{1 \times 2} + \frac{1}{2 \times 3} + \ldots + \frac{1}{L \times (L+1)} \\
&= \left(1 - \frac{1}{2}\right) + \left(\frac{1}{2} - \frac{1}{3}\right) + \ldots + \left(\frac{1}{L} - \frac{1}{L+1}\right) \\
&= 1 - \frac{1}{L+1}
\end{aligned} \quad (23)$$

We can further obtain:

$$\exp\left(-\sum_{j=1}^{L}\frac{1}{j+1}+\sum_{j=1}^{L}\frac{1}{(j+1)^2}\right)\|e_0\|^2$$
$$< \exp\left(-\ln\left(1+\frac{L}{2}\right)+1-\frac{1}{L+1}\right)\|e_0\|^2 \quad (24)$$
$$= \frac{2}{(L+2)}\exp\left(\frac{L}{L+1}\right)\|e_0\|^2$$

Finally, Eq. (20) can be formulated by:

$$\|e_L\|^2 \leq \frac{2}{(L+2)}\exp\left(\frac{L}{L+1}\right)\|e_0\|^2 \quad (25)$$

Hence, we can get $\lim_{L\to\infty}\|e_L\|=0$. This completes the proof.

The proposed OSCN (pseudo codes) is described in **OSCN Algorithm**.

---

**OSCN Algorithm**

Input: $X$ and output: $T$, $T_{max}$ as the maximum number of candidate nodes, $L_{max}$ as a maximum of hidden nodes, $\varepsilon$ as the training tolerance, $\sigma$ as a small positive number, Choose some positive scalars $\Upsilon = \{\lambda_{min}:\Delta\lambda:\lambda_{max}\}$

1. Initialization: $e_0 = T$ and $\Omega, W = [\ ], V_L = [\ ]; L=1$
2. **While** $L \leq L_{max}$ AND $\|e_0\| > \varepsilon$
  1). Hidden node Parameters Configuration (3-23)
3.   Set $r = L/(L+1)$
4.   **For** $\lambda \in \Upsilon$, **Do**
5.     **For** $i = 1,2,\cdots, T_{max}$, **Do**
6.       Randomly select candidate parameters ($w_L$, $b_L$) from $[-\lambda, \lambda]^d$ and $[-\lambda, \lambda]$
7.       Calculate $h_L$ based on (2), $\mu_L = (1-r)/(L+1)$;
8.       Calculate $v_L$: **if** $L=1$: $v_L = h_L$;
9.           **else**: Calculate $v_L$ based on (3)
10.       **If** $\|v_L\| \geq \sigma$
11.         Calculate $\xi_{L,q}$ based on (18)
12.         **If** $\min\{\xi_{L,1}, \xi_{L,2}, \cdots, \xi_{L,m}\} \geq 0$
13.           Save $w_L$, $b_L$ in $W$, $v_L$ in $V_L$, and $\xi_L$ in $\Omega$.
14.         End If
15.       **Else** go back to **Step 4**
16.       End If
17.     **End For** (corresponds to step 4)
18.     **If** $W$ is not Empty
19.       Find $w_L^*, b_L^*, v_L^*$ maximizing $\xi_L$ in $\Omega$
20.       Set $V = [v_1^*, v_2^*, \cdots, v_L^*]$
21.       **Break** (go to **Step 24**)
22.     **Else** randomly take $\tau = (1/2(1-r), 1-r)$ and let $r = r + \tau$, Return to **Step 4**;
23.     **End If**
24.   **End For** (corresponds to **Step 3**)
  2). Output Weight Determination
25.   Compute $\beta_L^* = \frac{\langle e_{L-1}, v_L^*\rangle}{\langle v_L^*, v_L^*\rangle}$;
26.   Compute $e_L = e_{L-1} - v_L^* \beta_L^*$;
27.   Update $e_0 := e_L$; $L = L+1$;
28. **End While**
29. **Return** $\beta_1^*, \beta_2^*, \cdots, \beta_L^*$, $w^*$ and $b^*$.

---

## 4 PERFORMANCE EVALUATION

In this section, in order to further substantiate the effectiveness and superiority of the proposed algorithms, comparisons among OSCN, SCN (here refer to SC-III), IRVFLN [24] are conducted through two numerical examples, twenty real-world regression and classification cases. We have selected real-world datasets deriving from the UCI database [35] and KEEL [36].

The activation function is denoted as $g(x) = 1/(1+\exp(-x))$ involving all the algorithms. According to the experience in [27], we set $r$ to 0.999 for SCN directly. The parameters $\tau_L(r + \mu_L)$ for OSCN are given according to **Theorem 2**. Besides, the parameter $\sigma$ is typically set to 1e-6 [37], but the value will be set to adjust to specific cases. The experiment averaged 50 trials. In addition, for each function approximation and benchmark dataset, and the Average (AVE) and Standard Deviations (DEV) of Root Mean Squares Error (RMSE) are displayed in the corresponding tables, respectively. For each classification case, we give the AVE and DEV in the corresponding tables as well.

The experiments concerning all algorithms are performed in MATLAB 2017a simulation software platform under a PC that configures with Intel Xeon W-2123, 3.6GHz CPU, 16G RAM.

### A. Regression cases

Two numerical examples, including single and multiple outputs, and ten real-world cases are investigated to evaluate the overall performance among OSCN, SCN and IRVFLN in regression cases within this part.

Numerical examples are the real-valued function generated by

$$y(x) = 0.2e^{-(10x-4)^2} + 0.5e^{-(80x-40)^2} + 0.3e^{-(80x-20)^2} \quad (26)$$

The dataset contains 1000 randomly generated samples from the uniform distribution [0, 1], among which 800 samples are chosen as the training set, whereas the testing set is comprised of 200 samples. For all testing algorithms, the expected training error tolerance $\varepsilon$ of RMSE is 0.05. In addition, the parameter $\sigma$ equals to 1e-6 for OSCN. The maximum number of candidate nodes $T_{max}$ and a maximum of hidden nodes $L_{max}$ of SCN-based algorithms are set to 20 and 100, respectively. $\Upsilon = \{150:10:200\}$ for SCN-based algorithms and [-150, 150] for IRVFLN. And the experimental results including the network architecture complexity, training time, AVE and DEV of training and testing RMSE with regard to all algorithms are reported in Table I.

As seen from Table I, in the case of achieving the same stop RMSE, OSCN requires fewer hidden nodes and takes about the same amount of time to train as SCN, and on these basis it still has smaller RMSE and DEV.

We set stop RMSE to 0.01 to compare these algorithms in approximation capability, the detailed convergence and fitting curves are plotted in Fig. 3, which shows the variation trend of RMSE accompanied by the increasing number of hidden nodes, and the corresponding approximation capability, respectively. For this function $y$, the results for IRVFLN are worse than others in both nodes and approximation. Compared to SCN, the algorithm we proposed not only achieves the same desired approximation, but also effectively reduces the number of iterations.

To further compare the performance of three algorithms, multiple-outputs numerical example is employed in this part. Two inputs $x_1$, $x_2$, two intermediate variables $x_3$, $x_4$ and two outputs $y_1$ and $y_2$ are used [38]. The relationships of this numerical example can be shown as followings:

$$\text{Inputs}: x_1, x_2$$
$$x_3 = x_1 + x_2$$
$$x_4 = x_1 - x_2 \quad (27)$$
$$\text{Outputs}: y_1 = \exp[2x_1 \sin(\pi x_4) + \sin(x_2 x_3)]$$
$$y_2 = \exp[2x_2 \cos(\pi x_3) + \cos(x_1 x_4)]$$

The inputs are randomly generated from $N(-0.5, 0.2)$, including 600 training data and 400 testing data whose means and variances are -0.5 and 0.2. The maximum times of random configuration $T_{max}$ and the scope are set to 10 and $\{10:5:50\}$, respectively. Moreover, the parameter $\sigma$ equals to 1e-8 for OSCN. In order to explore more differences among the three algorithms, the number of hidden nodes are setup for 4, 6 and 8, respectively. In the case of convergence of the entire model, results of training two outputs are shown in Table II.

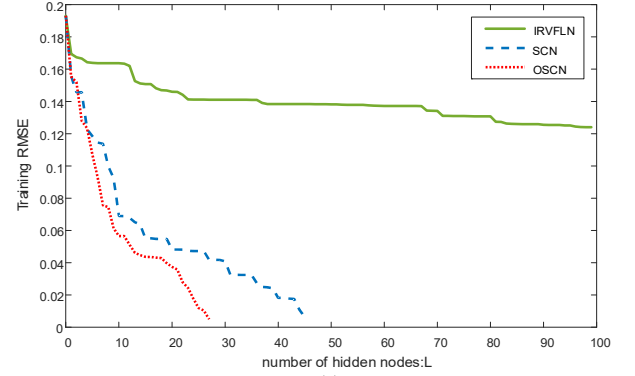

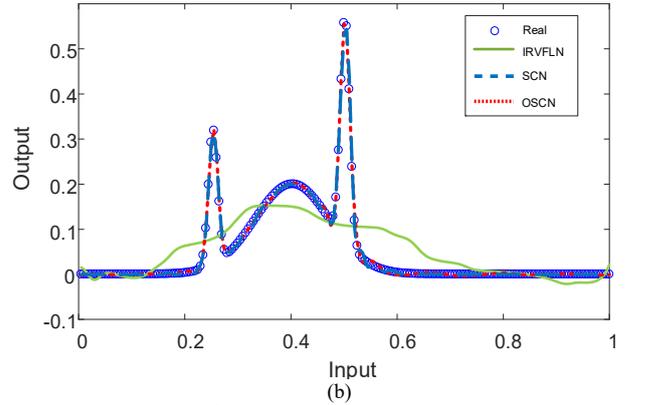

Fig. 3. Training results of three algorithms with $\varepsilon = 0.01$. (a) convergence curves. (b) fitting curves.

TABLE I
PERFORMANCE COMPARISONS ON FUNCTION $Y$.

| Algorithms | Stop RMSE ($\varepsilon$=0.05) | | | | | |
| --- | --- | --- | --- | --- | --- | --- |
| | Training | | | Testing | | |
| | $L$ | $t(s)$ | AVE | DEV | AVE | DEV |
| IRVFLN | 100 | 0.3601 | 0.1253 | 0.0062 | 0.1255 | 0.0062 |
| SCN | 25.46 | 0.0930 | 0.0438 | 0.0063 | 0.0437 | 0.0063 |
| OSCN | 15.75 | 0.1137 | 0.0429 | 0.0061 | 0.0428 | 0.0060 |

TABLE II
PERFORMANCE COMPARISON OF DIFFERENT NUMBER OF HIDDEN NODES.

| No. nodes | | 4 | | 6 | | 8 | |
| --- | --- | --- | --- | --- | --- | --- | --- |
| RMSE | | AVE | DEV | AVE | DEV | AVE | DEV |
| IRVFLN | $y_1$ | 0.3502 | 0.1202 | 0.3207 | 0.1071 | 0.3022 | 0.1127 |
| | $y_2$ | 0.3477 | **0.0263** | 0.3232 | **0.0307** | 0.3149 | 0.0315 |
| SCN | $y_1$ | 0.1442 | 0.0311 | 0.1221 | 0.0281 | 0.1047 | 0.0220 |
| | $y_2$ | 0.2561 | 0.0474 | 0.2050 | 0.0404 | 0.1688 | 0.0344 |
| OSCN | $y_1$ | **0.1426** | **0.0305** | **0.1113** | **0.0217** | **0.0897** | **0.0169** |
| | $y_2$ | **0.2215** | 0.0425 | **0.1692** | 0.0326 | **0.1273** | 0.0306 |

Comparisons are carried out from the perspective of AVE and DEV of the training RMSE on condition that all algorithms can achieve the same nodes. The convergent rate of OSCN, which outperforms SCN and IRVFLN, in each phase is apparent, especially in later phase of $y_2$. The appearances indicate that the added nodes of OSCN are more conducive to residual error decline. Moreover, the estimation variance (VAR) information of each data $(x_1, x_2)$ within 50 trials on $y_1$ and $y_2$ are given in Figs. 4-6, where the corresponding variance of each data is shown in the contour distribution. It can be clearly seen from Figs. 4-6, the variance distributions of SCN and IRVFLN are larger than those of OSCN. These experimental results, as displayed in Table II and Figs. 4-6, confirm that OSCN has the faster convergence rate and stronger stationary in training DEV and estimation VAR of each data for this numerical example compared with SCN and IRVFLN.

Finally, we illustrate the efficiency and feasibility of OSCN through more complex real-world regression cases. Employing ten real-world regression problems is to achieve the goal of performance evaluation. Meanwhile, the relevant information of real-world regression cases is displayed in Table III. Table IV shows detailed parameter settings including their expected training error tolerances $\varepsilon$ of RMSE. In addition, the parameter $\sigma$ equals to 1e-6 for OSCN.

Comparisons among OSCN, SCN, IRVFLN, corresponding to the number of hidden nodes and the AVE and DEV of training RMSE in which the predetermined error tolerance for each datasets is given, are drawn, as displayed in Table V. It worth mentioning that, with the same expected error, OSCN requires fewer network nodes than both IRVFLN and SCN in all real-world cases, even though SCN has the capability to achieve a relatively compact network through its supervisory mechanism.

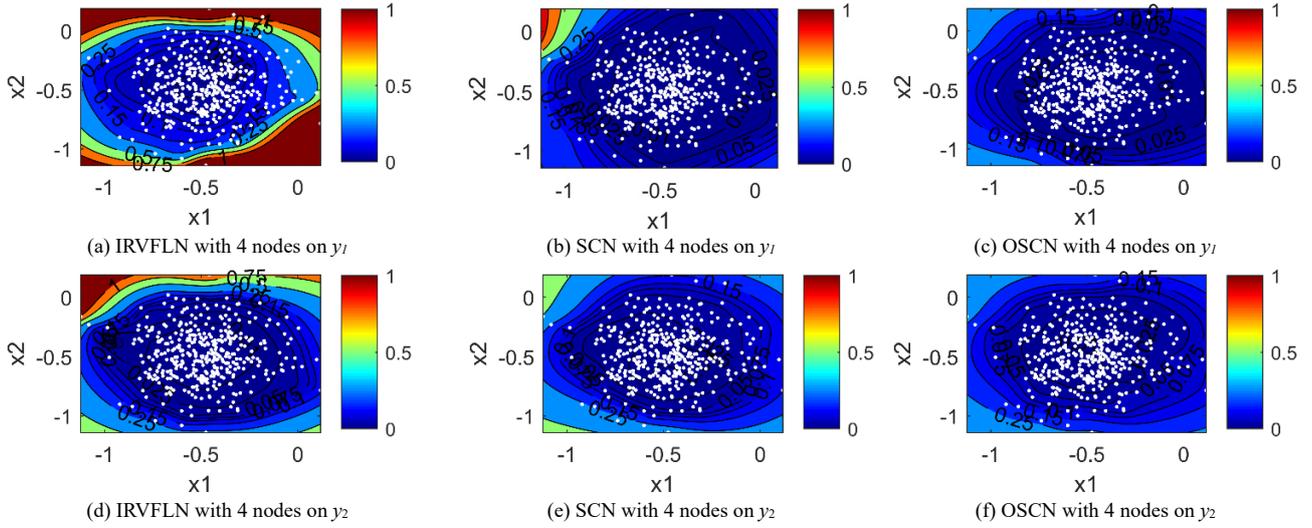

Fig. 4. Estimation variance using different learning models with 4 nodes on each data ($x_1$, $x_2$).

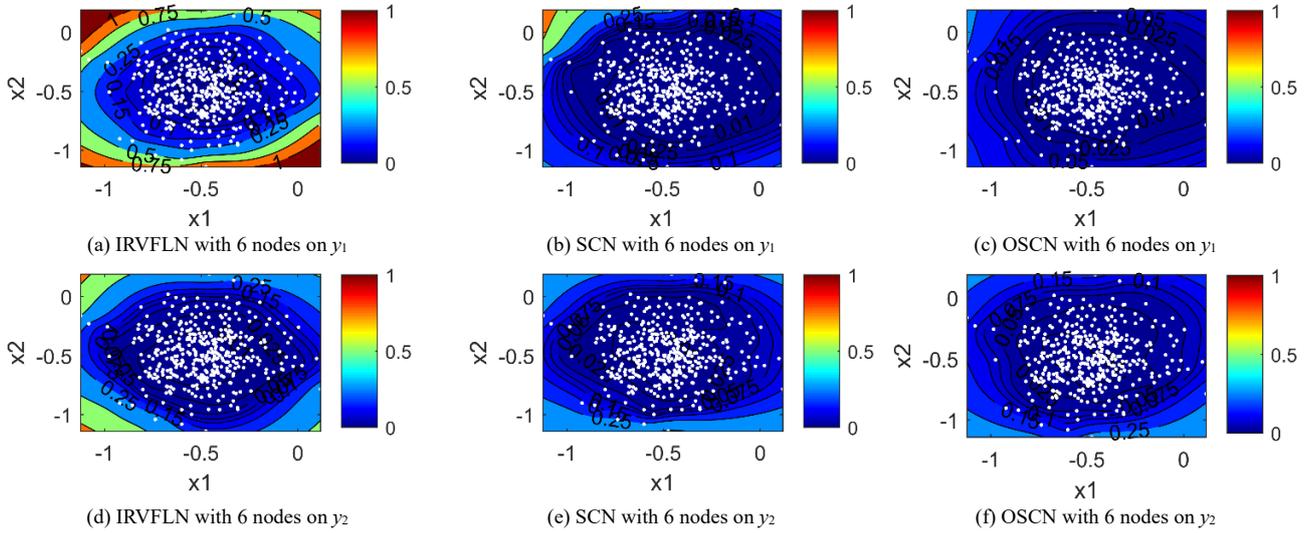

Fig. 5. Estimation variance using different learning models with 6 nodes on each data ($x_1$, $x_2$).

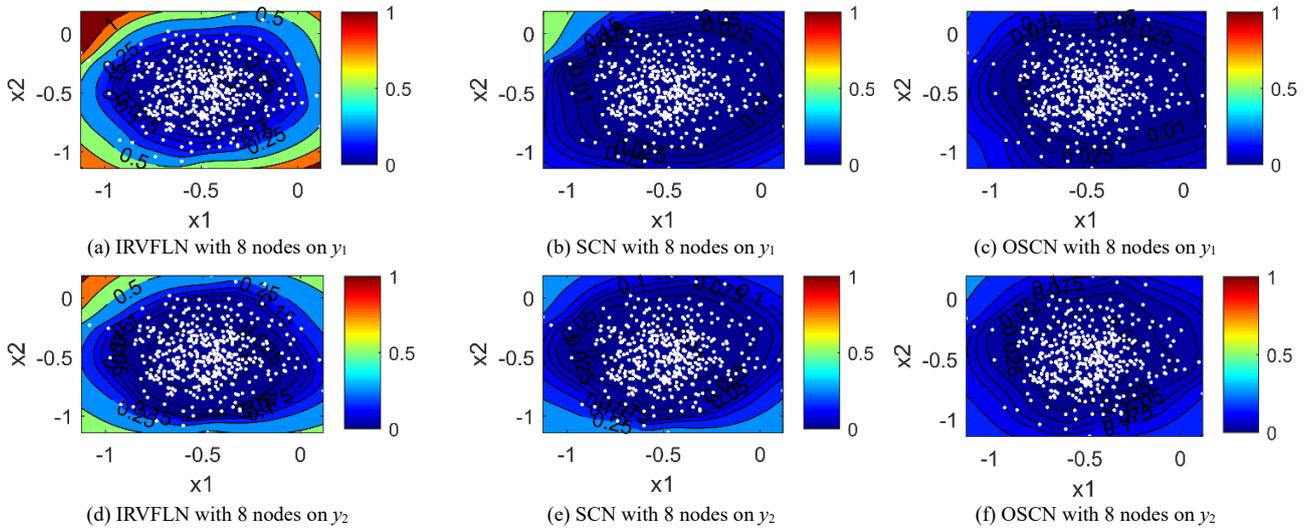

Fig. 6. Estimation variance using different learning models with 8 nodes on each data ($x_1$, $x_2$).

TABLE V
PERFORMANCE COMPARISONS OF TRAINING RMSE.

| Date sets | IRVFLN | | | SCN | | | OSCN | | |
|---|---|---|---|---|---|---|---|---|---|
| | Nodes | Training | | Nodes | Training | | Nodes | Training | |
| | | AVE | DEV | | AVE | DEV | | AVE | DEV |
| Abalone | 10 | 0.1234 | 0.0153 | 4.6 | 0.0886 | 0.0011 | **3.84** | **0.0885** | **0.0010** |
| Forestfire | 150 | 0.0712 | **7.46e-5** | 120.04 | **0.0598** | 0.0010 | 98.02 | 0.0599 | 0.0001 |
| Concrete | 250 | 0.1479 | 0.0059 | 172.66 | **0.0499** | **0.0001** | 156.64 | 0.0499 | 0.0001 |
| Winequality | 10 | 0.1949 | 0.0487 | 7.78 | 0.1295 | 0.0012 | **6.36** | **0.1292** | **0.0007** |
| Compactiv | 50 | 0.1941 | 0.0767 | 43.8 | 0.0494 | 0.0009 | **22.86** | **0.0493** | **0.0008** |
| Stock | 10 | 0.1478 | 0.0325 | 8.25 | 0.1054 | 0.0006 | **7.31** | **0.1047** | **0.0004** |
| Plastic | 200 | 0.1507 | 0.0033 | 162.48 | 0.0632 | 0.0014 | **132.72** | **0.0524** | **0.0002** |
| Pumadyn | 100 | 0.1390 | 0.0228 | 96.62 | **0.0622** | 0.0010 | 75.42 | 0.0772 | **0.0004** |
| Mortgage | 250 | 0.1254 | 0.0047 | 168.89 | 0.0478 | 0.0009 | **156** | **0.0485** | **0.0001** |
| Ankara | 10 | 0.0920 | 0.0151 | 7.56 | 0.0812 | **0.0010** | **4.2** | **0.0743** | 0.0012 |

TABLE VI
PERFORMANCE COMPARISONS OF TESTING RMSE.

| Date sets | IRVFLN | | SCN | | OSCN | |
|---|---|---|---|---|---|---|
| | AVE | DEV | AVE | DEV | AVE | DEV |
| Abalone | 0.1209 | 0.0152 | 0.0881 | 0.0027 | **0.0878** | **0.0026** |
| Forestfire | **0.0315** | 0.0408 | 0.1077 | 0.0252 | 0.0481 | **0.0012** |
| Concrete | 0.1629 | 0.0058 | 0.1169 | 0.0216 | **0.1029** | **0.0055** |
| Winequality | 0.2574 | 0.3712 | 0.1303 | 0.0044 | **0.1289** | **0.0012** |
| Compactiv | 0.2505 | 0.2085 | 0.0771 | 0.0905 | **0.0636** | **0.0415** |
| Stock | 0.2062 | 0.2334 | 0.1158 | 0.0982 | **0.1011** | **0.0930** |
| Plastic | 0.1825 | 0.1226 | 0.0847 | **0.0511** | **0.0768** | 0.0517 |
| Pumadyn | 0.1271 | 0.0325 | 0.0896 | 0.0025 | **0.0839** | **0.0012** |
| Mortgage | 0.1552 | 0.0079 | 0.0758 | 0.0036 | **0.0426** | **0.0023** |
| Ankara | 0.0924 | 0.0163 | 0.0809 | 0.0044 | **0.0779** | **0.0032** |

TABLE III
SPECIFICATION OF REAL-WORLD REGRESSION CASES.

| Data sets | No. of observation | | Attributes |
|---|---|---|---|
| | Training | Testing | |
| Abalone | 2000 | 2177 | 7 |
| Forestfire | 300 | 217 | 12 |
| Concrete | 772 | 258 | 8 |
| Winequality | 3428 | 1470 | 11 |
| Compactiv | 6144 | 2048 | 21 |
| Stock | 750 | 200 | 10 |
| Plastic | 1320 | 330 | 2 |
| Pumadyn | 6553 | 1639 | 32 |
| Mortgage | 839 | 210 | 15 |
| Ankara | 1287 | 322 | 9 |

TABLE IV
PARAMETER SETTINGS OF REAL-WORLD REGRESSION CASES.

| Data sets (Stop RMSE) | Algorithms parameters ($L_{max}, \Upsilon, T_{max}$) | | |
|---|---|---|---|
| | IRVFLN | SCN | OSCN |
| Abalone (0.09) | 10,{1},1 | 10,{1:0.5:10},10 | 10,{1:0.5:10},10 |
| Forestfire (0.06) | 150,{1},1 | 150,{1:0.5:10},10 | 150,{1:0.5:10},10 |
| Concrete (0.05) | 250,{1},1 | 250,{1:5:50},10 | 250,{1:5:50},10 |
| Winequality (0.13) | 10,{10},1 | 10,{10:5:50},10 | 10,{10:5:50},10 |
| Compactiv (0.05) | 50,{10},1 | 50,{10:1:20},10 | 50,{10:1:20},10 |
| Stock (0.11) | 10,{10},1 | 10,{1:0.5:10},10 | 10,{1:0.5:10},10 |
| Plastic (0.07) | 200,{10},1 | 200,{10:1:20},10 | 200,{10:1:20},10 |
| Pumadyn (0.09) | 100,{10},1 | 100,{10:1:20},10 | 100,{10:1:20},10 |
| Mortgage (0.05) | 250,{1},1 | 250,{1:5:50},10 | 250,{1:5:50},10 |
| Ankara (0.09) | 10,{1},1 | 10,{1:0.5:10},10 | 10,{1:0.5:10},10 |

Combining Table VI with Table V, we can see that OSCN and SCN have better approximate property than IRVFLN obviously under most circumstances. IRVFLN basically cannot reach our expect error under the specified $L_{max}$. OSCN is better at prediction performance and more compact than SCN under the similar AVE of training RMSE. Furthermore, OSCN exhibits exceptional advantages in the case of the overall testing RMSE, compared to SCN, especially in complex real-world situations such as Concrete and Compactiv, which indicates that OSCN may perform more favorably when it comes to complex datasets. The performance of OSCN on Forestfire data, however, is inferior to IRVFLN, but better than SCN.

### B. Real-world classification cases

In this second part, the classification performance of the proposed algorithm is validated in comparison of SCN and IRVFLN when it comes to the same number of hidden node. Ten selected datasets, which stem from real-world multiclass classification problems, are to make comparisons on training and testing accuracy. The relevant descriptions about them can be found in Table VII. Furthermore, the parameter settings of algorithms are shown in Table VIII. Table IX gives the results of comparison in the AVE and DEV of training and testing accuracy.

In general, the effect of the classification experiments is not as obvious as that of the regression experiments. But, as found in Table IX, OSCN is still much excellent in training accuracy and testing accuracy on the whole than both SCN and IRVFLN on condition that nodes is kept identical. As far as the Pima dataset is concerned, IRVFLN is significantly more stable than other two algorithms, but its accuracy is much lower than SCN and OSCN. Generally speaking, one has a preference for the expected accuracy than the most stable results with poor performance on expecting, thus OSCN and SCN are the better choices.

TABLE VII
SPECIFICATION OF REAL-WORLD CLASSIFICATION CASES.

| Data sets | No. of observation | | Attributes | Classes |
|---|---|---|---|---|
| | Training | Testing | | |
| Iris | 120 | 30 | 4 | 3 |
| Breast | 340 | 229 | 30 | 2 |
| Pima | 537 | 231 | 8 | 2 |
| Satimage | 4504 | 1931 | 36 | 6 |
| Page Blocks | 1315 | 800 | 10 | 5 |
| Banana | 3200 | 800 | 2 | 2 |
| Segment | 2079 | 231 | 19 | 7 |
| Vehicle | 716 | 80 | 18 | 4 |
| PenBased | 9490 | 1050 | 16 | 10 |
| Image segmentation | 1386 | 924 | 18 | 7 |

TABLE VIII
PARAMETER SETTINGS OF REAL-WORLD CLASSIFICATION CASES.

| Data sets ($L_{max}$) | Algorithms parameters ($\Upsilon$, $T_{max}$, $\sigma$) | | |
|---|---|---|---|
| | IRVFLN | SCN | OSCN |
| Iris (10) | {0.5} | {0.5:0.5:10},10 | {0.5:0.5:10},10,1e-6 |
| Breast (50) | {1} | {1:0.5:10},10 | {1:0.5:10},10,1e-4 |
| Pima (50) | {1} | {1: 0.5:50},10 | {1: 0.5:50},10, 1e-4 |
| Satimage (200) | {1} | {1:1:10},10 | {1:1:10},10, 1e-4 |
| Page Blocks (200) | {1} | {1:1:10},10 | {1:1:10},10, 1e-6 |
| Banana (150) | {1} | {1:1:10},10 | {1:1:10},10, 1e-6 |
| Segment (200) | {2} | {1:1:10},10 | {1:1:10},10, 1e-6 |
| Vehicle (100) | {1} | {1:1:10},10 | {1:1:10},10, 1e-6 |
| PenBased (300) | {1} | {1:1:10},10 | {1:1:10},10, 1e-6 |
| Image segmentation (200) | {1} | {1:1:10},10 | {1:1:10},10, 1e-6 |

TABLE IX
THE RESULTS OF COMPARISON OF TRAINING AND TESTING ACCURACY.

| Data sets | Algorithms | Training Accuracy | | Testing Accuracy | |
|---|---|---|---|---|---|
| | | AVE | DEV | AVE | DEV |
| Iris | IRVFLN | 0.4921 | 0.1838 | 0.4806 | 0.1732 |
| | SCN | **0.9823** | 0.0052 | 0.9360 | **0.0334** |
| | OSCN | 0.9805 | **0.0047** | **0.9413** | 0.0432 |
| Breast | IRVFLN | 0.9204 | 0.0198 | 0.9286 | 0.0223 |
| | SCN | 0.9898 | 0.0035 | 0.9549 | 0.0101 |
| | OSCN | **0.9928** | **0.0032** | **0.9593** | **0.0088** |
| Pima | IRVFLN | 0.6572 | 0.0043 | 0.6439 | 0.0068 |
| | SCN | 0.8102 | **0.0057** | 0.7677 | 0.0115 |
| | OSCN | **0.8140** | **0.0057** | **0.7720** | 0.0132 |
| Satimage | IRVFLN | 0.7626 | 0.0151 | 0.7795 | 0.0151 |
| | SCN | 0.9121 | **0.0018** | 0.8819 | 0.0036 |
| | OSCN | **0.9165** | 0.0026 | **0.8857** | **0.0032** |
| Page Blocks | IRVFLN | 0.9624 | 0.0027 | 0.8750 | 0.0150 |
| | SCN | 0.9673 | 0.0043 | 0.8862 | **0.0076** |
| | OSCN | **0.9722** | **0.0016** | **0.8869** | 0.0133 |
| Banana | IRVFLN | 0.8325 | 0.0036 | 0.7923 | 0.0152 |
| | SCN | 0.9000 | 0.0024 | 0.8963 | 0.0025 |
| | OSCN | **0.9006** | **0.0020** | **0.8978** | **0.0016** |
| Segment | IRVFLN | 0.9596 | **0.0018** | 0.8961 | 0.0064 |
| | SCN | 0.9702 | 0.0021 | 0.9134 | 0.0042 |
| | OSCN | **0.9822** | 0.0019 | **0.9394** | **0.0031** |
| Vehicle | IRVFLN | 0.8743 | 0.0091 | 0.7625 | 0.0347 |
| | SCN | 0.9018 | 0.0088 | 0.8455 | 0.0275 |
| | OSCN | **0.9134** | **0.0057** | **0.8750** | **0.0177** |
| PenBased | IRVFLN | 0.9934 | 0.0005 | 0.9923 | 0.0032 |
| | SCN | 0.9942 | 0.0005 | 0.9930 | 0.0024 |
| | OSCN | **0.9953** | **0.0002** | **0.9933** | **0.0015** |
| Image segmentation | IRVFLN | 0.7813 | 0.0068 | 0.7182 | 0.0067 |
| | SCN | 0.9773 | 0.0019 | 0.9498 | 0.0053 |
| | OSCN | **0.9800** | **0.0017** | **0.9528** | **0.0051** |

## 5 CONCLUSION

This paper proposes an advanced learning approach for SCNs with orthogonalization technology. The proposed orthogonal SCN (OSCN) can avoid generating redundant nodes to reduce the complexity of network and enhance the convergence performance. Concretely, OSCN makes the candidate nodes orthogonal to the existing nodes, and abandons poor candidates according to a criterion of node quality. Then, an orthogonal form of supervisory mechanism is established to guarantee the universal approximation property. Under the framework of OSCN, the global optimal parameters can be determined analytically by employing an incremental updating scheme, which is reported in detail in this paper. Theories and experimental results illustrate that OSCN performs better performance on reducing the number of convergence iterations, improving stability, as well as approximation and estimation capacities. The future work will explore the proposed approach in combination with block increments to further reduce the number of iterations while improving modeling efficiency.

## REFERENCES

[1] G. Deshpande, P. Wang, D. Rangaprakash, and B. Wilamowski, "Fully connected cascade artificial neural network architecture for attention deficit hyperactivity disorder classification from functional magnetic resonance imaging data," *IEEE Trans. Cybern.*, vol. 45, no. 12, pp. 2668-2679, Dec. 2015.

[2] W. Dai, Q. Liu, T.-Y. Chai, "Particle size estimate of grinding processes using random vector functional link networks with improved robustness," *Neurocomputing*, vol. 169, pp. 361-372, 2015.

[3] S.-C. Huang and B.-H. Do, "Radial basis function based neural network for motion detection in dynamic scenes," *IEEE Trans. Cybern.*, vol. 44, no. 1, pp. 114-125, Jan. 2014.

[4] N. Najmaei and M. R. Kermani, "Applications of Artificial Intelligence in Safe Human-Robot Interactions," *IEEE Trans. Syst., Man, and Cybern. B, Cybern.*, vol. 41, no. 2, pp. 448-459, April. 2011.

[5] K. Dai, J. Zhao, F. Cao, "A novel algorithm of extended neural networks for image recognition," *Eng. Appl. Artif. Intel.*, vol. 42, pp. 57-66, Jun. 2015.

[6] L. Ma, K. Khorasani, "Facial expression recognition using constructive feedforward neural networks," *IEEE Trans. Syst., Man, and Cybern. B, Cybern.*, vol. 34, no. 3, pp. 1588-1595, Jun. 2004.

[7] M. Wang, W. Fu, X. He, S. Hao and X. Wu, "A survey on large-scale machine learning," *IEEE Trans. Knowl. Data Eng.*, vol. 34, no. 6, pp. 2574-2594, June. 2022.

[8] F. Ahmad, A. Abbasi, B. Kitchens, D. Adjeroh and D. Zeng, "Deep learning for adverse event detection from web search," *IEEE Trans Knowl. Data Eng.*, vol. 34, no. 6, pp. 2681-2695, June. 2022.

[9] M. Alhamdoosh, D. Wang. "Fast decorrelated neural network ensembles with random weights," *Information Sciences*, vol. 264, pp. 104-117, Apr. 2014.

[10] S. Lin, J. Zeng and X. Zhang, "Constructive neural network learning," *IEEE Trans. Cybern.*, vol. 49, no. 1, pp. 221-232, Jan. 2019.

[11] F. Cao, D. Wang, H. Zhu, "An iterative learning algorithm for feedforward neural networks with random weights," *Information Sciences*, vol. 1, no. 9, pp. 546-557, 2016.

[12] W. F. Schmidt, M. A. Kraaijveld and R. P. W. Duin, "Feedforward neural networks with random weights," *Proceedings, 11th IAPR Int. Conf. Pattern Recognition. Vol.II. Conference B: Pattern Recognition Methodology and Systems*, The Hague, Netherlands, 1992, pp. 1-4.

[13] S. Scardapane, D. Wang, Randomness in neural networks: an overview, *Wiley Interdisciplinary Reviews Data Mining & Knowledge Discovery*, vol. 7, no. 2, e1200, Mar./Apr. 2017.

[14] X.-Z. Wang, T. Zhang, R. Wang, "Noniterative Deep Learning: Incorporating Restricted Boltzmann Machine Into Multilayer Random Weight Neural Networks," *IEEE Trans. Syst., Man, and Cybern.: Syst*, vol. 49, no. 7, pp. 1299-1308, Jul. 2017.

[15] C. L. P. Chen, J. Z. Wan, "A rapid learning and dynamic stepwise updating algorithm for flat neural networks and the application to time-series prediction," *IEEE Trans. Syst., Man, and Cybern. B, Cybern*, vol.29, no. 1, pp. 62-72, Feb. 1999.

[16] B. Igelnik and Y.-H. Pao, "Stochastic choice of basis functions in adaptive function approximation and the functional-link net," *IEEE Trans. Neural Netw.*, vol. 6, no. 6, pp. 1320-1329, Nov. 1995.

[17] Y.-H. Pao, G.-H. Park, and D. J. Sobajic, "Learning and generalization characteristics of the random vector functional-link net," *Neurocomputing.*, vol. 6, no. 2, pp. 163-180, 1994.

[18] Y.-H. Pao and Y. Takefuji, "Functional-link net computing, theory, system architecture, and functionalities," *IEEE Comput*., vol. 3, no. 5, pp. 76-79, May. 1992.

[19] K. Xu, H. Li, H. Yang, "Kernel-Based Random Vector Functional-Link Network for Fast Learning of Spatiotemporal Dynamic Processes," *IEEE Trans. Syst., Man and Cybern., Syst.*, vol. 49, no. 5, pp. 1016-1026, May 2019.

[20] H. Ye, F. Cao, D. Wang, "A hybrid regularization approach for random vector functional-link networks," *Expert Systems With Applications*, vol. 140, pp. 12912, 2020.

[21] S. Scardapane, D. Wang and A. Uncini, "Bayesian random vector functional-link networks for robust data modeling," *IEEE Trans. Cybern.*, vol. 48, no. 7, pp. 2049-2059, July. 2018.

[22] L. Zhang and P. N. Suganthan, "Visual tracking with convolutional random vector functional link network," *IEEE Trans. Cybern.*, vol. 47, no. 10, pp. 3243-3253, Oct. 2017.

[23] T.-Y. Kwok and D.-Y. Yeung, "Constructive algorithms for structure learning in feedforward neural networks for regression problems," *IEEE Trans. Neural Netw.*, vol. 8, no. 3, pp. 630-645, May 1997.

[24] T.-Y. Kwok and D.-Y. Yeung, "Objective functions for training new hidden units in constructive neural networks," *IEEE Trans. Neural Netw*, vol. 8, no. 8, pp. 1131-1148, 1997.

[25] M. Li, D.-H. Wang, "Insights into randomized algorithms for neural networks: practical issues and common pitfalls," *Information Sciences*, vol. 382-383, pp. 170-178, Mar. 2017.

[26] A. N. Gorban, I. Y. Tyukin, D. V. Prokhorov, and K. I. Sofeikov, "Approximation with random bases: Pro et contra," *Information Sciences*, vols. 364-365, pp. 129-145, Oct. 2016.

[27] D. Wang and M. Li, "Stochastic Configuration Networks: Fundamentals and Algorithms," *IEEE Trans. Cybern.*, vol. 47, no. 10, pp. 3466-3479, Oct. 2017.

[28] J. Lu, J. Ding, "Construction of prediction intervals for carbon residual of crude oil based on deep stochastic configuration networks," *Information Sciences*, vol. 486, pp. 119-132, 2019.

[29] D. Wang, M. Li, "Deep stochastic configuration networks with universal approximation property", arXiv:1702.0563918, 2017, to appear in IJCNN, 2018.

[30] J. Lu and J. Ding, "Mixed-Distribution Based Robust Stochastic Configuration Networks for Prediction Interval Construction," *IEEE Trans. Ind. Inform.*, vol. 16, no. 8, pp. 5099-5109, Aug. 2020.

[31] J. Lu, J. Ding, X. Dai and T. Chai, "Ensemble stochastic configuration networks for estimating prediction intervals: a simultaneous robust training algorithm and its application," *IEEE Trans. Neural Netw. Learn Syst.*.vol. 31, no. 12, pp. 5426-5440, Dec. 2020.

[32] M. Li, D. Wang, "2-D Stochastic Configuration Networks for Image Data Analytics," *IEEE Trans. Cybern.*, vol. 51, no. 1, pp. 359-372, Jan. 2021.

[33] P. Lancaster, M. Tismenetsky, The Theory of Matrices: With Applications, Amsterdam, Netherlands: Elsevier, 1985.

[34] T. S. Shores, Applied linear algebra and matrix analysis: 2nd ed.. Gewerbestrasse, Switzerland: Springer International Publishing AG, part of Springer Nature, 2018, pp. 206-312.

[35] C. L. Blake, C. J. Merz. "UCI repository of machine learning databases," Dept. Inf. Comput. Sci., Univ. California, Irvine, CA, 1998. [Online]. Available: http://www.ics.uci.edu/~mlearn/MLRepository.html.

[36] J. Alcalá-Fdez, A. Fernández, J. Luengo, J. Derrac, S. García, L. Sánchez, F. Herrera, "Keel data-mining software tool: data set repository, integration of algorithms and experimental analysis framework," *J. Mult. Valued Log. Soft Comput*, vol. 17, no. 2-3, pp. 255-287, 2011.

[37] P. Zhou, Y. Jiang, C. Wen, T.-Y. Chai, "Data modeling for quality prediction using improved orthogonal incremental random vector functional-link networks," *Neurocomputing*, vol. 365, pp. 1-9, 2019.

[38] Y. Liu, Q.-Y. Wu, J. Chen, "Active selection of informative data for sequential quality enhancement of soft sensor models with latent variables," *Industrial & Engineering Chemistry Research*. Vol. 56, no. 16, pp. 4804-4817, 2017.